\documentclass[10pt]{article}

\usepackage[letterpaper,textwidth=6in,textheight=8.9in,centering]{geometry}
\usepackage[T1]{fontenc}
\usepackage[tt=false,type1=true]{libertine}
\usepackage[varqu]{zi4}
\usepackage[libertine]{newtxmath}
\usepackage{microtype}

\usepackage{titlesec}
\titleformat{\section}{\sffamily\bfseries\large}{\thesection}{0.75em}{\MakeUppercase}
\titleformat{\subsection}{\sffamily\bfseries\normalsize}{\thesubsection}{0.75em}{}
\titleformat{\subsubsection}{\sffamily\itshape\normalsize}{\thesubsubsection}{0.75em}{}
\titlespacing*{\section}{0pt}{2.2ex plus 1ex minus .2ex}{1ex plus .2ex}
\titlespacing*{\subsection}{0pt}{1.8ex plus 1ex minus .2ex}{0.8ex plus .2ex}

\usepackage[labelfont={sf,bf},font={sf,small},labelsep=period,skip=6pt]{caption}
\usepackage{graphicx}
\usepackage{amsmath}
\usepackage{booktabs}
\usepackage{tabularx}
\usepackage{enumitem}
\usepackage[numbers,sort&compress,square]{natbib}
\usepackage{xcolor}
\definecolor{ACMPurple}{cmyk}{0.55,1,0,0.15}
\definecolor{ACMDarkBlue}{cmyk}{1,0.58,0,0.21}
\usepackage[colorlinks,linkcolor=ACMPurple,citecolor=ACMPurple,urlcolor=ACMDarkBlue,bookmarksnumbered,unicode]{hyperref}
\usepackage{cleveref}

\newcommand{\doi}[1]{doi: \href{https://doi.org/#1}{#1}}

\newcommand{\figfile}[1]{%
  \IfFileExists{#1.png}%
    {\includegraphics[width=\linewidth]{#1.png}}%
    {\fbox{\parbox[c][0.25\textheight][c]{0.9\linewidth}{\centering\texttt{#1}}}}}

\makeatletter
\renewcommand{\maketitle}{%
  \begingroup\raggedright
  {\sffamily\bfseries\LARGE\@title\par}\vskip 1.2em
  {\sffamily\large\@author\par}\vskip 0.3em
  {\small\@date\par}\vskip 1.5em
  \endgroup}
\makeatother
\renewenvironment{abstract}{\small\noindent{\sffamily\bfseries Abstract.}\ }{\par\vskip 1em}

\title{CktHGVAE: A Symmetric Design/State Heterogeneous Graph Variational Autoencoder for Transistor-level Analog/Mixed-signal Circuit Design and Analysis}
\author{Shigeyasu Uno}
\date{Ritsumeikan University, 1-1-1 Noji-Higashi, Kusatsu, Shiga 525-8577, Japan\\
  \texttt{suno@fc.ritsumei.ac.jp}}

\begin{document}
\maketitle

\begin{abstract}
We propose CktHGVAE, a heterogeneous graph variational autoencoder for transistor-level analog/mixed-signal CMOS circuits that takes not only design variables but also state variables as its inputs and outputs.
With a single hyperparameter configuration without topology-specific tuning, comparison between model-generated and SPICE-simulated circuit operational states gives median $R^2$ values above 0.9 for typical single-ended voltage amplifiers and the CMOS inverter, and 0.81-0.96 for differential amplifiers with larger datasets.
Latent-space optimization in the PCA-reduced latent space with Bayesian optimization, evolutionary search, random search, and quasi-exhaustive search is demonstrated for a common-source voltage amplifier with NMOS input and resistive load, where the objective function is defined as a penalty for deviation from required circuit specifications.
For each search strategy, 93.3 - 100\% of all design candidates suggested by the model satisfy the required circuit specifications at the decoder output, and SPICE simulations verify that 90 - 100\% of them indeed satisfy the requirements.
The model pretrained on a batch of topologies can be transferred to unseen topologies without retraining at all, with zero-shot reconstruction median $R^2$ values of 0.72-0.90 in three of four scenarios, including transfer from single-ended to differential topologies, although the transfer fidelity depends on the choice of pretraining topologies.
Our demonstrations suggest that the proposed model can potentially serve as a basis for AMS CMOS circuit design automation.

\end{abstract}

\section{Introduction}\label{SectionIntroduction}

As the need for semiconductor chips grows in wide range of industries, electronic circuit designing based on complementary metal-oxide-semiconductor (CMOS) large-scale integrated (LSI) technology has been gaining much more importance.
While digital circuit design has already been highly automated, analog and mixed signal (AMS) circuit design automation still needs significant advancement \cite{NooriZadeh2025,Guven2025}.
One of the reasons is that the analog circuit performances are highly dependent on subtle differences in transistor operating conditions, process-specific characteristics and variabilities.
Therefore, in determining design parameters such as transistor sizes and values of passive elements (resistance/capacitance/inductance) in electronic circuits, AMS CMOS circuit designers depend heavily on their physical insights, experiences, and many trials of numerical simulations by SPICE simulators.

Efforts to bring automation in determining design parameters have mainly been directed toward optimization of design variables: given an objective function based on design requirements, find a set (vector) of design parameters minimizing/maximizing the objective function \cite{Gao2023,Guven2025}.
Some of such efforts were found being successful in finding optimal transistor sizes for some important analog circuit topologies.
However, such optimization in high-dimensional design space often suffer from the `curse of dimensionality', being limited to several tens of design parameters \cite{Gao2023,Chen2022}.
Also, it still needs many trials of circuit simulations until reaching the final optimum.

In recent years, utilization of deep learning has emerged as a new direction of AMS CMOS circuit design automation which could overcome above challenges in optimization methods.
Some works use multi-layer perceptron (MLP) as a surrogate model of circuit simulator: for a given set of design parameters, it outputs key performance indicators (KPI) \cite{Rashid2024}.
Other works use variational autoencoder (VAE) neural network \cite{Kingma2014} for dimensionality reduction in design parameters: VAE constructs low-dimensional latent space to represent input design variables, and optimization of objective function is performed in this reduced-dimensional space \cite{Li2024}.
The VAE is also used to augment training dataset to reduce actual number of SPICE simulations \cite{Poddar2024}.
Other forms of generative models have also been used such as large-language-models (LLM) \cite{Chen2025,Lai2025,Chang2024,Yu2026,Ahmadzadeh2025}, transformer \cite{Gao2025}, and combination of transformer and graph representation \cite{Pathak2026,Kim2026}.

Many of state-of-the-art machine learning models for AMS CMOS circuit design automation incorporate the graph-neural-networks (GNN) \cite{Wu2021}, which is in fact a natural and reasonable path because electronic circuits are inherently graphs of device connection.
FALCON uses GNN to construct a surrogate model of circuit simulator: for a given topology and design parameters, GNN is used to calculate target KPIs \cite{Mehradfar2025}.
While optimal circuit topology is found by a feed-forward classification MLP, transistor sizing is done by using this GNN-based surrogate model and gradient-based optimization.
OPAMP-Generator also uses GNN to design operational amplifier end-to-end in multiple steps \cite{Lu2023}.
First, circuit topology is represented as directed acyclic graph (DAGs) of amplifier elements with macromodel description to encode both topology and element characteristics using graph variational autoencoder (GVAE).
Optimal topology and element characteristics are then found by bi-level Bayesian optimization in the latent space, followed by device sizing based on gm/Id methodology.
CktGNN treats a whole operational amplifier as DAG of connection between amplifying elements represented by macromodels \cite{Dong2023}.
It then uses gated recurrent unit (GRU) based aggregation function for message passing in encoding, and the whole architecture is based on VAE.
CktGen treats a whole amplifier consisting of several amplifying elements represented by macromodels similarly to CktGNN, and it encodes the whole circuit adjacency matrix using GNN as well as element type and position vectors \cite{Hou2026}.
These are used as inputs of transformer encoder to construct latent space, and transformer decoder is conditioned by circuit specifications (design requirements) to generate both circuit topology and element characteristics which meet required specifications.
The whole architecture is basically conditional VAE (CVAE) with latent space alignment by contrastive learning.

These previous works indicate several important implications in leveraging neural networks for AMS CMOS circuit design automations.
First, GNN is a natural and effective representation of electronic circuit operation \cite{Mehradfar2025,Lu2023,Dong2023,Hou2026}.
Second, VAE is a preferred architecture thanks to its ability in reduced-order latent space optimization \cite{Li2024}, generation of (or augmentation to) unseen circuit operation/dataset/topologies \cite{Poddar2024,Dong2023,Hou2026}, and conditional circuit generation for circuit design automation and optimization workflow \cite{Hou2026}.
Also, generation of latent space and analysis of its structure could potentially lead to providing a `meaning' and `explanation' of a given circuit operation, which is highly helpful for AMS circuit designers but not available from conventional black-box optimization methods.
However, state-of-the-art models such as CktGNN and CktGen mainly focus on topology generation with circuit blocks represented by macromodels, so their latent spaces are not suitable for transistor-level modeling of circuit behaviour \cite{Dong2023,Hou2026}.
Device-level graph representations have also been used \cite{Gao2025,Kim2026}, but mainly for generating circuit topologies rather than for modeling the operating state of a given circuit.
In terms of latent space construction, the VAE-based approaches reviewed above use design variables such as device types and transistor sizes (and circuit topology) as inputs/outputs of the VAE.
Performance information enters the latent space only indirectly, for example through specification conditioning as in CktGen \cite{Hou2026}, and KPIs or state variables such as node voltages and currents need to be derived by a separate neural network or by circuit simulation \cite{Li2024}.
Therefore, the latent space constructed in this way represents the design space only, and does not directly encode the circuit operation (state variables).

Now, if we take both design and state variables as inputs/outputs of the VAE, the latent space would represent the joint manifold of design and state variables obtained from actual circuit simulation results.
Such symmetric treatment of design/state variables eliminates the need for a separate neural network or SPICE simulations to derive KPIs determined by the operating point, since the state variables output by the VAE can be used to calculate them.
To the best of our knowledge, no GVAE that treats design and state variables symmetrically as its inputs and outputs has been proposed for AMS CMOS circuit design.

In this work, we propose a simple but effective neural network model architecture to solve the issues described above, named CktHGVAE: circuit heterogeneous graph variational autoencoder.
Its architecture is as simple as graph variational autoencoder where both encoder/decoder consist of graph neural networks applied to circuit topological graph.
Both inputs/outputs consist of a mixture of design variables (transistor sizes, device types, applied voltages, etc.) and state variables (node voltages, currents, etc.) mapped on each node of the topological graph.
Here we demonstrate the following:
\begin{enumerate}[label=(\alph*),nosep]
  \item our model can successfully learn the DC operating-point circuit operation of 12 fundamental circuit topologies, tested against circuit simulation results
  \item our model provides specification-directed optimization search in the PCA-reduced latent space with Bayesian optimization, evolutionary search, random search, and quasi-exhaustive search, whose suggested designs are verified by SPICE simulations
  \item our model pretrained on a group of topology is well transferred to unseen topology, providing `zero-shot' cross-topology transfer.
\end{enumerate}

\section{Method}\label{SectionMethod}

\subsection{Neural Network Model Architecture}\label{SubSectionMethodModelArchitecture}

\Cref{FigArchitechtureOverview} outlines the structure of CktHGVAE, which is based on the conventional VAE structure with GNN as encoder and decoder.
It processes a batch of electronic circuits in various operating conditions.
The variables associated with each device/net are categorized into design and state variables: `design variables' refer to those which act as inputs of circuit simulator (voltage/current source values, transistor sizes, passive element values, etc.), while `state variables' refer to outputs from the simulator (voltages, currents, transconductance, etc.).
Circuit-specific global characteristics that are determined by the DC operating point, such as bias currents, gm/ID, and low-frequency small-signal voltage gain, can be analytically calculated from the simulated state variables, or can be optionally assigned to a device as additional state variables.
The framework used in this work for DC operating-point analysis can be extended to AC analysis with only marginal extension, for example by including capacitive state variables and small-signal simulation results in the graph.

\begin{figure}[tbp]
  \centering
  \figfile{FigArchitechtureOverview}
  \caption{Schematic overview of CktHGVAE architecture. Electronic circuit is viewed as a collection of topology, constants, design variables, and state variables. The graph builder converts the circuit topology into a heterogeneous graph, and preprocessor assigns constants and variables according to SPICE simulation results. Encoder takes it to construct latent variables, and those variables are used to sample latent variables to be passed to the decoder, and the decoder reconstructs the original variables. In training phase, the input to the decoder will be generated by the reparameterization trick, while in design candidate generation phase the latent variables are directly passed to the decoder. During training, comparing the reconstructed variables with the original ones, the loss function is calculated and used to update learnable parameters in encoder/decoder. For circuit design automation, the latent space is fitted by PCA and GMM, and optimization search is performed in the PCA-reduced latent space.}
  \label{FigArchitechtureOverview}
\end{figure}

First, the graph builder converts the circuit topology into a graph with multiple types of nodes and edges, also known as heterogeneous graph (hetero-graph).
Nodes represent different types of devices (n/p-type transistors and other passive elements), pins (drain/gate/source/bulk for transistors, positive/negative for passive elements), and nets (circuit connection points).
This graph conversion style is basically the same as that of AnalogGenie \cite{Gao2025}, but we directly process the graph using GNN.
Edges represent connections between those different types of nodes, and each edge is identified by what types of nodes it is connecting.
As some of the variables assigned to the nodes could vary by orders of magnitude, the preprocessor takes absolute, scale, and log of some of the variables.

Such variable-assigned hetero-graph is then taken as input of the encoder, which converts the scaled physical variables to hidden variables by MLP heads specific to node types.
Then, message passing is calculated by GraphSAGE convolution as follows \cite{Hamilton2017}:
\begin{equation}
  \mathbf{h}_v' = \mathbf{W}_{\mathrm{self}} \cdot \mathbf{h}_v + \mathbf{W}_{\mathrm{neighbor}} \cdot \mathrm{AGG}(\{\mathbf{h}_u : u \in N(v)\}),
\end{equation}
where $\mathbf{h}_v$ and $\mathbf{h}_v'$ are hidden variables before/after the message passing, $N(v)$ is the set of source-side neighbors connected to a node $v$, $\mathrm{AGG}(*)$ is aggregation function (taking sum in this work), $\mathbf{W}_{\mathrm{self}}$ and $\mathbf{W}_{\mathrm{neighbor}}$ refer to learnable linear mapping matrices dependent on different edge types, respectively.
This message passing layer is repeated several times to realize message propagation over circuit topology structure.
Resulting hidden variables are finally converted to mean ($\mu$) and log-variance (and hence standard deviation $\sigma$) of latent variables for each node by MLP heads.
It is important to note that the encoder learns (a) how to convert physical variables to hidden variables for each node type, (b) how to pass messages for each edge type and each message passing layer, and (c) how to convert resulting hidden variables to latent variable $\mu$'s and $\sigma$'s for each node type.
Therefore, the learnable parameters are `topology-agnostic', meaning that CktHGVAE can take training datasets consisting of either single topology or multiple different topologies, enabling `zero-shot' cross topology transfer demonstrated in Section~\ref{SubSectionResultsCrossTopologyTransfer}.
The decoder does exactly the reverse of the encoder: it takes sampled latent variables for each node, calculates message passing, then applies the per-node-type MLP heads to output physical values assigned to each node.

During training, $\mu$'s and $\sigma$'s for each latent variable are used to sample a set of latent variables by the reparameterization trick and passed to the decoder \cite{Kingma2014}.
Then, the loss function is calculated as follows:
\begin{equation}
  L_{\mathrm{loss}} = L_{\mathrm{recon}} + \beta L_{\mathrm{kld}},
\end{equation}
where $L_{\mathrm{recon}}$ represents the reconstruction (mean square) error calculated from input to the encoder and output from the decoder, $L_{\mathrm{kld}}$ is KL divergence term, and the coefficient $\beta$ is varied according to beta-annealing method \cite{Higgins2017}.
The learnable parameters in the encoder/decoder are determined by minimizing $L_{\mathrm{loss}}$ after several training epochs.
Details of hyperparameters are shown in \Cref{AppendixHyperparameters}.

\subsection{Latent Space Fitting and Search}\label{SubSectionMethodLSO}

The concatenation of latent variables for every node constitutes the latent space associated with the circuit topology and design/state variables (circuit operations), and holds the manifold of physical phenomena within the circuit.
However, as the number of nodes for a given topology is much larger than actual degree of freedom (DOF) of the circuit operation, this latent space is over-expressive and its dimensions can be further reduced without appreciable loss of meaningful information.
Such dimension reduction contributes to efficient search for optimal design for a given required circuit specifications in circuit design automation, and also physical interpretation will be much easier than original latent space.
In this work, we employ the principal component analysis (PCA) to find major axes of latent space manifold, and then use gaussian mixture model (GMM) to fit probabilistic function within the PCA-reduced latent space.

Circuit design automation using CktHGVAE will be performed as latent-space optimization (LSO): it is formulated as finding optimal design variables (transistor sizes, passive element values, input voltages, etc.) which minimizes a given objective function representing penalty out of circuit specifications.
The objective function is defined as
\begin{align}
  \mathrm{objective}(\mathbf{p}) &= \mathrm{spec\_score}(\mathbf{p}) + \mathrm{density\_penalty}(\mathbf{p}), \\
  \mathrm{spec\_score}(\mathbf{p}) &= \sum_i \mathrm{weight}_i \cdot \max(0, \mathrm{rel}_i(\mathbf{p}) - \mathrm{tolerance}), \\
  \mathrm{rel}_i(\mathbf{p}) &=
  \begin{cases}
    (\mathrm{target}_i - \mathrm{value}_i(\mathbf{p})) / |\mathrm{target}_i| & \text{for conditional description ``>=''}\\
    (\mathrm{value}_i(\mathbf{p}) - \mathrm{target}_i) / |\mathrm{target}_i| & \text{for ``<=''}\\
    |\mathrm{value}_i(\mathbf{p}) - \mathrm{target}_i| / |\mathrm{target}_i| & \text{for ``==''},
  \end{cases} \\
  \mathrm{density\_penalty}(\mathbf{p}) &= \mathrm{penalty\_weight} \cdot \max(0, \mathrm{min\_percentile} - \mathrm{density\_percentile}(\mathbf{p})),
\end{align}
where $\mathrm{spec\_score}(\mathbf{p})$ penalizes deviation from physical variable specifications described using $i$th target value ($\mathrm{target}_i$) and its conditions (``>='', ``<='', ``==''), and $\mathrm{value}_i(\mathbf{p})$ represents the $i$th physical variable decoded (via the decoder) and rescaled (via postprocessor) from a PCA-reduced latent vector $\mathbf{p}$.
The tolerance is a small relative slack (set as 0.001: 0.1\%) below which a deviation is not penalized, and it is applied to all conditions, and is necessary for the `==' condition to be satisfiable.
The function $\mathrm{density\_penalty}(\mathbf{p})$ penalizes the latent space optimization search to pull search trials within the range of not-too-small probability density regions.
Here, $\mathrm{density\_percentile}(\mathbf{p})$ is the percentile rank of the GMM-fitted probability density at $\mathbf{p}$ among those of the training data, and $\mathrm{min\_percentile}$ is the threshold below which the penalty applies.
Throughout this paper, min\_percentile = 10.0 and penalty\_weight = 1.0 will be used.
Note that the objective function value is always greater or equal to zero: its value is zero when all the requirements are satisfied.

In this study we consider four major search strategies:
\begin{enumerate}[label=(\alph*),nosep]
  \item Bayesian optimization (BO): a typical conventional black-box optimization search.
  \item Covariance matrix adaptation evolution strategy (CMA-ES) search: an evolutionary search that adapts a multivariate Gaussian search distribution \cite{Hansen2023}.
  \item Random search (RND): multiple candidates are randomly sampled in PCA-reduced space according to the probability density fitted by GMM, and sorted in objective-function-value order to find the top candidates.
  \item Quasi-exhaustive search (QES): multiple candidates are quasi-evenly sampled by Sobol sampling \cite{Sobol1967} within a box of the PCA-reduced space (the population mean $\pm$ 4 standard deviations along each principal component), regardless of the GMM-fitted probability, and sorted in objective function value order; low-probability regions are discouraged only through the density penalty in the objective function.
\end{enumerate}

\subsection{Closed-loop Fidelity Verification}\label{SubSectionMethodFidelityVerification}

For a given pretrained model and datasets, the model will construct full latent space and associated PCA-reduced latent space.
The design automation by LSO explained in Sec.~\ref{SubSectionMethodLSO} will be successful if and only if input/output physical variables and latent space are all consistent.
This can be checked in the following procedure:
\begin{enumerate}[label=(\arabic*),nosep]
  \item Pretrain the model (determine learnable parameters in encoder/decoder) using a dataset.
  \item Construct the full latent space, find PCA-reduced latent space and fit GMM using the same dataset.
  \item Draw new random points in the PCA-reduced latent space, inverse-PCA-transform to find the full latent space vectors, and decode them to find physical (design/state) variables for each graph nodes.
  \item Use the design variables from the generated samples to run SPICE simulations to compare the state variables (transistor operation conditions, net voltages, currents, etc.) between model-generated and SPICE-simulated ones.
\end{enumerate}
The final comparison accuracy is evaluated using $R^2$ values.

\subsection{Zero-shot Cross-topology Transfer Capability Test}\label{SubSectionMethodCrossTopologyTransfer}

As CktHGVAE employs GNN in encoder/decoder, its learnable parameters are topology-agnostic, as explained in \Cref{SubSectionMethodModelArchitecture}.
Therefore, it can be trained on a set of topologies, and this pretrained model can be used for unseen (held-out) circuit topologies without retraining its learnable parameters.
A SPICE-simulated dataset of the held-out topology is needed only to fit its latent-space model.
This ability is checked in the following procedure:
\begin{enumerate}[label=(\arabic*),nosep]
  \item Prepare two datasets: one dataset (denoted as DataP) including various circuit topologies and operation conditions, the other dataset (denoted as DataH) including a held-out circuit topology which is not included in the topology lists in DataP.
  DataH is split into train/validation/test splits (70/15/15\%).
  \item Pretrain the model using the DataP to determine learnable weights within the encoder/decoder, and freeze those weights for the rest of this procedure (therefore, no retraining).
  \item Using the pretrained model, encode the training split of DataH with the frozen encoder, and fit the PCA (and GMM) to the resulting latent vectors (this step is not retraining, just a fitting with little computational burden).
  \item Encode each graph in the test split of DataH, project its latent vector onto the first $k$ principal components, transform it back to the full latent space, and decode it.
  The decoded physical values are compared with the original input values of the same graph, and reconstruction $R^2$ is calculated for each physical quantity (design and state variables).
\end{enumerate}

Steps (2)-(4) are repeated for five pretraining random seeds \{42, 101, 202, 303, 404\}, while the train/validation/test splits of DataP and DataH kept identical across seeds.
The mean and standard deviation of the resulting $R^2$ over the seeds are then reported.
The number of principal components $k$ is determined by the same explained-variance rule used for the latent-density fitting for the training split of DataH, and the same $k$ is used for all pretraining seeds so that the seeds are compared at an identical latent dimensionality.

Note that this procedure evaluates reconstruction of test graphs through the reduced latent space, not generation of new samples verified by SPICE simulation as in \Cref{SubSectionMethodFidelityVerification}.
Therefore the resulting $R^2$ values are not directly comparable with those in \Cref{SubSectionResultsFidelityVerification}.

\subsection{Circuit Topologies, Simulation and Dataset Preparation}\label{SubSectionMethodCircuitDatasets}

\begin{figure}[htbp]
  \centering
  \figfile{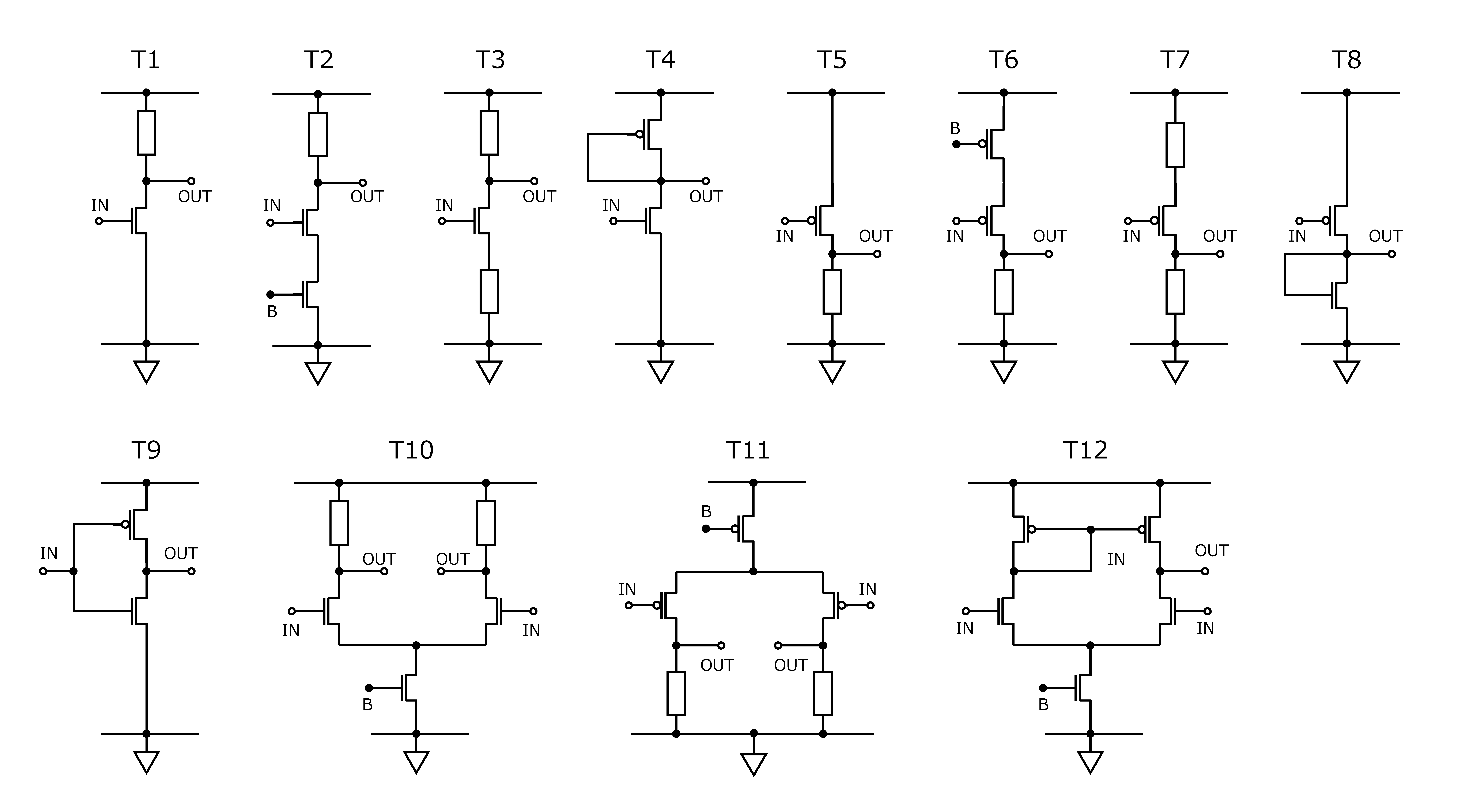}
  \caption{Circuit diagrams of T1-T12 topologies studied in this paper.
  Top/bottom rails correspond to VDD/GND, respectively, resistors are indicated by rectangles, and NMOS/PMOS are indicated using the conventional symbols, where NMOS bulk connections to GND and PMOS well connections to VDD are implicit (actually connected but not shown in these diagrams).
  IN/OUT are input/output nets, and B is the bias voltage net.}
  \label{FigCircuitTopologies}
\end{figure}

The circuit topologies tested in this study are listed in \Cref{TableCircuitTopologyList}, and actual circuit diagrams are shown in \Cref{FigCircuitTopologies}.
For each topology, SPICE operating-point (.op) analysis was done using Ngspice \cite{Ngspice}.
The SkyWater 130nm open-source process design kit (PDK), with its sky130\_fd\_pr\_{nfet, pfet}\_01v8 device models at the typical process corner (tt) and the default temperature of 27 °C, was used to provide realistic CMOS characteristics \cite{SkyWater}.
For each topology, multiple set of transistor widths, lengths, resistor values, input/bias voltages were sampled from specified ranges by Sobol sampling, while the supply voltage was kept constant at 1.8 V.
Transistor width/length ranges were [min: 0.18, max: 10.0, step: 0.02] um, and resistance value ranges [min: 1000, max: 5000, step: 10] Ohm.
The input voltage applied to the gate of input transistors ranges [min: 0.1, max: 1.7, step: 0.01] V.
The bias voltage applied to the current-source transistors ranges similar to that of input voltage, but limited to make sure that the current source transistors are always on.
For each topology, 1,024 Sobol samples were initially simulated (2,048 for the 2,000-sample datasets), and samples whose simulation failed were discarded, giving approximately 1,000 (2,000) data entries.
These data are divided into train/validation/test splits (70/15/15\%), where train/validation splits are used during model training, and the PCA and GMM are fitted on the train split, while test splits are preserved for other purposes whenever necessary.

\begin{table}[tbp]
  \caption{Circuit topologies tested in this paper. Labels, circuit name abbreviations, and descriptions are listed.}
  \label{TableCircuitTopologyList}
  \small
  \begin{tabularx}{\linewidth}{llX}
    \toprule
    Label & Abbreviation & Description \\
    \midrule
    T1  & CSA-N-Res      & Common-source voltage amplifier, NMOS input, resistive load \\
    T2  & CSA-N-Res-SDCS & Common-source voltage amplifier, NMOS input, resistive load, source-degenerated by NMOS current-source \\
    T3  & CSA-N-Res-SDRS & Common-source voltage amplifier, NMOS input, resistive load, source-degenerated by a resistor \\
    T4  & CSA-N-Diode-P  & Common-source voltage amplifier, NMOS input, diode-connected PMOS load \\
    T5  & CSA-P-Res      & Common-source voltage amplifier, PMOS input, resistive load \\
    T6  & CSA-P-Res-SDCS & Common-source voltage amplifier, PMOS input, resistive load, source-degenerated by PMOS current-source \\
    T7  & CSA-P-Res-SDRS & Common-source voltage amplifier, PMOS input, resistive load, source-degenerated by a resistor \\
    T8  & CSA-P-Diode-N  & Common-source voltage amplifier, PMOS input, diode-connected NMOS load \\
    T9  & CMOS-INV       & CMOS inverter (one n-type and one p-type transistor) \\
    T10 & DP-N-Res       & Differential-pair voltage amplifier, NMOS input pairs, resistive loads, NMOS current-source bias \\
    T11 & DP-P-Res       & Differential-pair voltage amplifier, PMOS input pairs, resistive loads, PMOS current-source bias \\
    T12 & DP-N-OTA       & Differential-pair voltage amplifier, NMOS input pairs, active current mirror loads, NMOS current-source bias (a 5-transistor operational transconductance amplifier) \\
    \bottomrule
  \end{tabularx}
\end{table}

\subsection{Software versions}\label{SubSectionMethodSoftware}
The model was implemented by Python 3.14, PyTorch 2.13.0 \cite{Paszke2019}, PyTorch Geometric 2.8 \cite{Fey2019}, numpy 2.5.1, scipy 1.18.0.
The PCA and GMM were implemented by scikit-learn 1.9.0 \cite{Pedregosa2011}, and BO and CMA-ES were implemented using Optuna 4.9.0 \cite{Akiba2019}.
The SPICE simulations were done using Ngspice 46.
All numerical computations were performed on MacBook Pro 2023 edition (macOS 26.6.2).

\section{Results}\label{SectionResults}

\subsection{Closed-loop Fidelity Verification}\label{SubSectionResultsFidelityVerification}

\begin{figure}[tbp]
  \centering
  \figfile{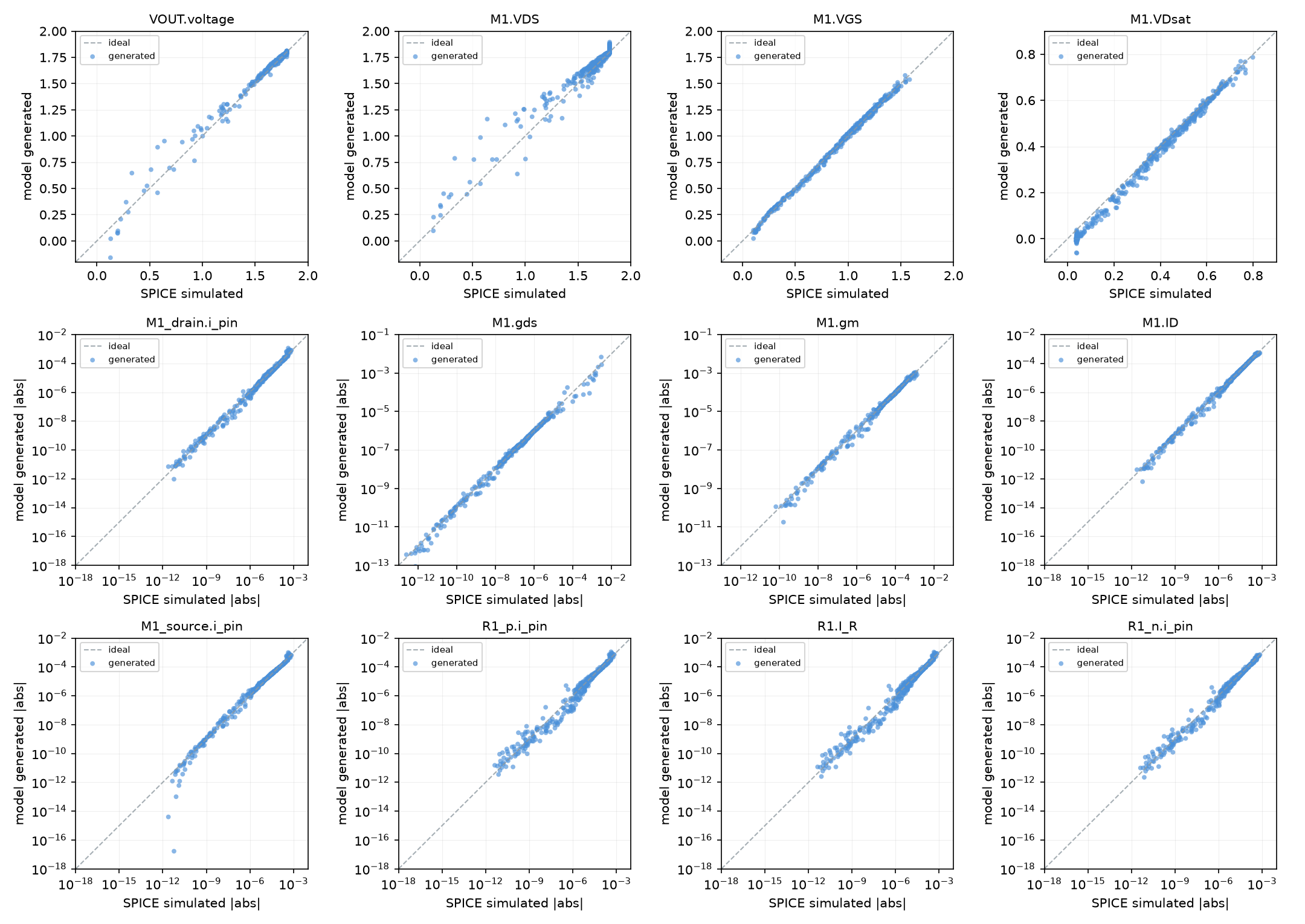}
  \caption{Comparisons between SPICE-simulated and model-generated values for state variables in CSA-N-Res (T1) circuit.
  M1 and R1 refer to the input NMOS and resistor, respectively.
  The VDS, VGS, VDsat, gds, gm, ID after `M1.' refer to state variables assigned to the NMOS, while \texttt{M1\_\{drain, source\}.i\_pin} refer to currents assigned to drain/source pins.
  The \texttt{R1.I\_R}, \texttt{R1\_p.i\_pin}, and \texttt{R1\_n.i\_pin} are currents assigned to resistor device and positive/negative terminal pins, respectively.
  \texttt{VOUT.voltage} refers to voltage assigned to output net.}
  \label{FigT1VerificationScatterPlots}
\end{figure}

\Cref{FigT1VerificationScatterPlots} shows comparisons between model-generated and SPICE-simulated state variables for CSA-N-Res (T1) circuit.
The symbols represent results from random samples which have design variables falling within the predefined value ranges for model training explained in Sec.~\ref{SubSectionMethodCircuitDatasets}.
Note that the overall agreement with generated and simulated values is reasonably good, demonstrating high fidelity of the trained model and constructed latent space manifold.

\Cref{TableT1FidelityVerification} shows the $R^2$ values of the verification results, where mean (and standard deviation) values from 5 different random seeds for PCA-reduced latent space sampling are shown for each state variables.
The $R^2$ values are calculated from in-domain samples only: samples whose design variables fall outside the predefined value ranges are excluded, and 2,189 out of 2,500 generated samples (87.6\%, 500 per seed) were in-domain.
The good agreement is persistent for different random seeds, and overall mean $R^2$ values are all well above 0.9.

\begin{table}[tbp]
  \caption{Mean (standard deviation) $R^2$ values for each state variables from 5 different random seeds for random sampling in PCA-reduced latent space, calculated from in-domain samples only.
  For each seed numbers \{42, 101, 202, 303, 404\}, number of samples with design variables within predefined ranges were \{430, 447, 438, 433, 441\} out of 500 trials, respectively (87.6\% in total).
  'Domain' indicates whether the state variables are treated in linear or log10 scale, and the $R^2$ values are calculated accordingly.
  }
  \label{TableT1FidelityVerification}
  \small
  \begin{tabular}{llc}
    \toprule
    Variable & Domain & Mean $R^2$ \\
    \midrule
    \texttt{VOUT.voltage}      & linear & 0.9731 ($\pm$ 0.0063) \\
    \texttt{M1.VDS}            & linear & 0.9250 ($\pm$ 0.0133) \\
    \texttt{M1.VGS}            & linear & 0.9979 ($\pm$ 0.0001) \\
    \texttt{M1.VDsat}          & linear & 0.9830 ($\pm$ 0.0011) \\
    \texttt{M1\_drain.i\_pin}  & log10  & 0.9946 ($\pm$ 0.0008) \\
    \texttt{M1.gds}            & log10  & 0.9932 ($\pm$ 0.0006) \\
    \texttt{M1.gm}             & log10  & 0.9950 ($\pm$ 0.0006) \\
    \texttt{M1.ID}             & log10  & 0.9962 ($\pm$ 0.0006) \\
    \texttt{M1\_source.i\_pin} & log10  & 0.9811 ($\pm$ 0.0129) \\
    \texttt{R1\_p.i\_pin}      & log10  & 0.9718 ($\pm$ 0.0040) \\
    \texttt{R1.I\_R}           & log10  & 0.9770 ($\pm$ 0.0030) \\
    \texttt{R1\_n.i\_pin}      & log10  & 0.9843 ($\pm$ 0.0018) \\
    \bottomrule
  \end{tabular}
\end{table}

\Cref{TableAllTopologyFidelityVerification} shows the closed-loop fidelity verification results for different circuit topologies, where $R^2$ values were again evaluated for in-domain samples.
The median and mean values over all state variables are presented, and their standard deviations over numbers of different random seeds are indicated in parentheses.
The rate of in-domain samples over all generated samples (500 per seed, pooled across seeds) is also listed in the table: it ranges from 36.3\% (T12, 1,000 samples) to 93.8\% (T5), and is generally lower for the differential pairs (T10-T12, 36.3-76.3\%) than for the single-ended amplifiers and the inverter (T1-T9, 74.8-93.8\%).
It is important to note that evaluations were performed with a single fixed hyperparameter setting, demonstrating how well CktHGVAE handles various circuit topologies without any topology-specific tuning.
Even under such modest setting, typical single-ended voltage amplifiers (T1 - T8) and CMOS inverter (T9) achieve median $R^2$ values better than 0.9, showing that the model captures the circuit operation of these topologies very well.
For differential pairs (T10-T12), median $R^2$ values are relatively weak (0.6908 - 0.9207) for approximately the same dataset size (around 1,000).
However, when the number of samples is doubled (around 2,000), the median $R^2$ improves for all three (0.8141 - 0.9561), and the mean $R^2$ and the in-domain rate also improve for T10 and T12.
One plausible explanation for this behavior is that the differential pairs involve more design variables (devices and nets) than the single-ended amplifiers, and therefore need more samples to cover the manifold of circuit operation.
In all of those topologies, the mean $R^2$ values are slightly lower than median $R^2$, indicating that some of the state variables show lower fidelity.
These results may be improved by more elaborate, topology-specific hyperparameter tuning whenever higher accuracy is required for specific design optimization needs.

\begin{table}[tbp]
  \caption{
    Median and mean $R^2$ over all state variables, averaged (with standard deviation) over different random seeds.
    Calculated from in-domain samples only.
    Approximate dataset size and in-domain sample rates (in-domain samples over all generated samples, 500 per seed, pooled across seeds) are also shown.
  }
  \label{TableAllTopologyFidelityVerification}
  \resizebox{\linewidth}{!}{%
  \begin{tabular}{llcccc r}
    \toprule
    Topology & Abbreviation & \shortstack{Approximate\\dataset size} & Median $R^2$ & Mean $R^2$ & \shortstack{Number of\\different seeds} & \shortstack{In-domain\\sample rate} \\
    \midrule
    T1  & CSA-N-Res      & 1,000  & 0.9858 ($\pm$ 0.0031) & 0.9810 ($\pm$ 0.0022) & 5 & 87.6\% (2,189/2,500) \\
    T2  & CSA-N-Res-SDCS & 1,000  & 0.9112 ($\pm$ 0.0101) & 0.8981 ($\pm$ 0.0098) & 8 & 78.9\% (3,157/4,000) \\
    T3  & CSA-N-Res-SDRS & 1,000  & 0.9959 ($\pm$ 0.0003) & 0.9900 ($\pm$ 0.0003) & 5 & 87.3\% (2,182/2,500) \\
    T4  & CSA-N-Diode-P  & 1,000  & 0.9912 ($\pm$ 0.0005) & 0.9893 ($\pm$ 0.0004) & 5 & 80.0\% (2,001/2,500) \\
    T5  & CSA-P-Res      & 1,000  & 0.9856 ($\pm$ 0.0011) & 0.9722 ($\pm$ 0.0056) & 5 & 93.8\% (2,346/2,500) \\
    T6  & CSA-P-Res-SDCS & 1,000  & 0.9527 ($\pm$ 0.0032) & 0.8194 ($\pm$ 0.0194) & 8 & 78.5\% (3,138/4,000) \\
    T7  & CSA-P-Res-SDRS & 1,000  & 0.9784 ($\pm$ 0.0018) & 0.7492 ($\pm$ 0.0777) & 8 & 83.1\% (3,323/4,000) \\
    T8  & CSA-P-Diode-N  & 1,000  & 0.9827 ($\pm$ 0.0009) & 0.9793 ($\pm$ 0.0014) & 5 & 74.8\% (1,869/2,500) \\
    T9  & CMOS-INV       & 1,000  & 0.9190 ($\pm$ 0.0146) & 0.9015 ($\pm$ 0.0098) & 8 & 90.5\% (3,619/4,000) \\
    T10 & DP-N-Res       & 1,000  & 0.6908 ($\pm$ 0.0258) & 0.6455 ($\pm$ 0.0204) & 5 & 52.6\% (1,316/2,500) \\
    T10 & DP-N-Res       & 2,000  & 0.9197 ($\pm$ 0.0152) & 0.9130 ($\pm$ 0.0093) & 8 & 71.2\% (2,850/4,000) \\
    T11 & DP-P-Res       & 1,000  & 0.9207 ($\pm$ 0.0079) & 0.8256 ($\pm$ 0.0138) & 5 & 76.3\% (1,908/2,500) \\
    T11 & DP-P-Res       & 2,000  & 0.9561 ($\pm$ 0.0058) & 0.7697 ($\pm$ 0.0328) & 8 & 62.5\% (2,500/4,000) \\
    T12 & DP-N-OTA       & 1,000  & 0.7080 ($\pm$ 0.0292) & 0.5470 ($\pm$ 0.0310) & 5 & 36.3\% (908/2,500) \\
    T12 & DP-N-OTA       & 2,000  & 0.8141 ($\pm$ 0.0355) & 0.7731 ($\pm$ 0.0307) & 8 & 48.2\% (1,930/4,000) \\
    \bottomrule
  \end{tabular}}
\end{table}

\subsection{Latent Space Optimization}\label{SubSectionResultsT1LSO}

Here we demonstrate LSO using the CSA-N-Res (T1) circuit with the requirements for circuit specifications shown in \Cref{TableLSORequiredSpecs}.
As the objective function hits its minimum value (zero) whenever every constraint is satisfied, LSO search results may give multiple candidates of optimal designs for given requirements.
Therefore, rather than looking closely into one of many candidates, it is more appropriate to verify whether those suggested candidates from LSO search actually satisfy the requirements by running SPICE simulations.

\begin{table}[tbp]
  \caption{
    Circuit specifications used in latent space optimization of T1 circuit.
  }
  \label{TableLSORequiredSpecs}
  \small
  \begin{tabular}{llc}
    \toprule
    Quantity & Constraint & Weight \\
    \midrule
    \texttt{VOUT.voltage} & >= 1.45 V   & 1.0 \\
    \texttt{VOUT.voltage} & <= 1.65 V   & 1.0 \\
    \texttt{M1.ID}        & <= 2.0e-4 A & 1.0 \\
    \texttt{M1.gm}        & >= 1.0e-4 S & 2.0 \\
    \texttt{M1.gm/ID}     & >= 2.0      & 1.0 \\
    \bottomrule
  \end{tabular}
\end{table}

\Cref{TableT1LSOResults} shows such verification results for four different search strategies mentioned in \Cref{SubSectionMethodLSO}.
For BO and ES, single search run with 300 maximum iteration budget will give one candidate, and such run was repeated 30 times with different initial conditions, resulting in 30 best candidates.
Each run starts with 10 warm start trials drawn from the GMM-fitted probability with random seed for each run.
The best candidate is the earliest trial that reaches the minimum objective value.
For RND, single search run will pick one best candidate out of 1024 random draws in PCA-reduced latent space subject to GMM-fitted probability, and such run was repeated 30 times with different random seeds, resulting in 30 best candidates.
For QES, the same procedure as RND was taken, but 1024 candidates are drawn by Sobol sampling regardless of GMM-fitted probability.
For 30 candidates for each, decoded circuit operations are tested against the objective function to see if required specifications are satisfied (denoted as `satisfied at decoder output').
Finally, those passed candidates were tested against actual SPICE simulation to see if they truly satisfy the required specifications (denoted as `satisfied at SPICE simulations').

For all four strategies, majority of candidates suggested by the model were verified by SPICE simulation as truly satisfying the required circuit specifications: 28 out of 30 (93.3\%) for BO and ES, 30 out of 30 for RND, and 27 out of 28 for QES.
This suggests that LSO based on our model can provide useful candidates in design automation on a designer's request, at least for requirements of this level of difficulty.
The QES was the only strategy for which not every candidate satisfied the specifications at decoder output (28 out of 30).
As explained in \Cref{SubSectionMethodLSO}, QES draws candidates in the PCA-reduced latent space range within $\pm$ 4 standard deviations along each principal component, and therefore may visit regions where the model accuracy is lower, resulting in some candidates not satisfying the specifications at decoder output.
Similarly for RND, some candidates could be drawn from low-probability regions of the latent space.
In fact, the out-of-training-range candidates were found for RND (3 out of 30) and QES (2 out of 30).

The differences among the four strategies are small compared with the variation expected between repeated sets of 30 trials.
Such small differences do not necessarily reflect actual capability ranking among them, because the numbers could be affected by randomness or initial warm state trials.
In particular, since BO and ES start from the same warm start trials and the objective function reaches its minimum value (zero) as soon as all constraints are satisfied, a search run often returns one of these warm start trials.
With the requirements used here, the specifications can therefore be met without extensive search.
The RND and QES, on the other hand, evaluate their candidates in a single batch without sequential search, making them suitable for generating diverse exploratory candidates.

\begin{table}[tbp]
  \caption{Verification of LSO optimal design candidates based on requirements in \Cref{TableLSORequiredSpecs} compared among different search strategies.
  Each strategy was run for 30 independent search trials with different random seeds (initial search conditions for BO/ES).
  Satisfied at decoder output: of the 30 trials, number whose decoded design already satisfies every constraint within the tolerance of 1e-3.
  Satisfied at SPICE simulations: of those decoder-satisfied candidates, number that still satisfy every constraint after independent SPICE re-simulation.
  Out-of-training-range candidates: of the 30 trials, number whose decoded design variables fall outside the training dataset's value range for at least one input.}
  \label{TableT1LSOResults}
  \resizebox{\linewidth}{!}{%
  \begin{tabular}{lccc}
    \toprule
    Strategy & \shortstack{Number of candidates\\satisfied at\\decoder output} & \shortstack{Number of candidates\\satisfied at\\SPICE simulations} & \shortstack{Number of\\out-of-training-range\\candidates} \\
    \midrule
    Bayesian optimization (BO)    & 30 & 28 & 0 \\
    CMA-ES search (ES)            & 30 & 28 & 0 \\
    Random search (RND)           & 30 & 30 & 3 \\
    Quasi-exhaustive search (QES) & 28 & 27 & 2 \\
    \bottomrule
  \end{tabular}}
\end{table}

\subsection{Zero-shot Cross-topology Transfer}\label{SubSectionResultsCrossTopologyTransfer}

As mentioned in \Cref{SubSectionMethodCrossTopologyTransfer}, CktHGVAE's learnable parameters are topology-agnostic: once it is pretrained on wide variety of topologies, it could generalize to circuit operations of unseen topologies.
In order to demonstrate such generalization capability, we tested several different scenarios shown in \Cref{TableZeroShotCrossTopologyTransfer}.
Details of the verification procedure is described in \Cref{SubSectionMethodCrossTopologyTransfer}.

First of all, for scenarios B-D, the zero-shot reconstruction median $R^2$ values are 0.72-0.90, which are quite high considering that the model was pretrained on different topologies and never saw the held-out topologies during pretraining at all.
This is a strong evidence that our model can learn the underlying physics of circuit operations, and it can generalize to unseen topologies based on the learned representations.
The exception is scenario A, where the median $R^2$ value is close to zero, indicating that the model failed to generalize to the held-out topology T6.

Looking in more detail on scenario A, the model was pretrained on a batch of topologies T1 - T3, T5, and T7, and then the reconstruction $R^2$ value is calculated for the test split of T6.
Among the pretraining topologies, T1, T3, T5, T7 involve one NMOS/PMOS input device and resistors in source/drain pins of transistors, while T2 has two NMOS in series.
As the held-out topology T6 is basically a PMOS counterpart of T2 involving PMOS in series, the model might have applied the operation of the NMOS series connection learned from T2 to the PMOS series connection of T6.

In Scenario B, the pretraining topologies T4, T8, T9 are basically NMOS and PMOS in series connection, and such structure takes a major part in held-out topology T12 except for NMOS current source in the tail.
In this case, the model would have learned from pretraining how to treat PMOS/NMOS series connection, leading to a relatively high mean $R^2$ value of 0.720.
It is notable that a model pretrained only on single-ended topologies reconstructs the differential structure with a tail current source (T12) reasonably well.
This indicates that the model successfully learned the underlying physics of the series connection of NMOS/PMOS devices, and extended its understanding for differential structures.

For scenarios C and D, the model is pretrained on all single-ended topologies (T1 - T9: common-source voltage amplifiers and the CMOS inverter), while the held-out topologies are differential-pair voltage amplifiers.
In this case, the pretraining topologies involve NMOS/PMOS/resistor series connections in wide variety of topologies, which might have been enough ingredients to understand resistive-loaded NMOS/PMOS differential voltage amplifiers T10-T11.

\begin{table}[tbp]
  \caption{
    Zero-shot cross-topology transfer results compared among different scenarios.
    Reconstruction median $R^2$ of the held-out topology's test split (see \Cref{SubSectionMethodCrossTopologyTransfer}), shown as mean (standard deviation) over five pretraining random seeds \{42, 101, 202, 303, 404\}, with the test split fixed across seeds.
    The number of principal components $k$ is fixed by the explained-variance rule of the latent-density fitting applied to the seed-42 model: $k = \{4, 10, 12, 12\}$ for scenarios A-D.
    Scenarios C and D share the same pretrained models. 
  }
  \label{TableZeroShotCrossTopologyTransfer}
  \resizebox{\linewidth}{!}{%
  \begin{tabular}{cllcc}
    \toprule
    Scenarios & Pretrained on & Held out & \shortstack{Number of graphs\\in test split} & \shortstack{Zero-shot reconstruction\\median $R^2$} \\
    \midrule
    A & T1, T2, T3, T5, T7 & T6  & 293 & 0.007 ($\pm$ 0.149) \\
    B & T4, T8, T9         & T12 & 279 & 0.720 ($\pm$ 0.100) \\
    C & T1 - T9            & T10 & 292 & 0.901 ($\pm$ 0.019) \\
    D & T1 - T9            & T11 & 287 & 0.824 ($\pm$ 0.059) \\
    \bottomrule
  \end{tabular}}
\end{table}

\section{Discussion}\label{SectionDiscussion}

The capabilities of our model demonstrated in \Cref{SectionResults} are considered to stem from its architecture.
Symmetric treatment of design and state variables as input/output of a straightforward VAE architecture removes the need for additional MLP branches for circuit operational state or KPI prediction.
The PCA-reduced latent space manifold construction leads to a flexible and easy-to-use design optimization framework, as well as future physical explainability of circuit operations.
The GNN-based implementation of the encoder/decoder enables the zero-shot cross-topology transfer, although the reconstruction fidelity varied among the scenarios.
These basic characterizations and demonstrations suggest a range of possible use cases, which will be explored in our future work.

It is tempting to impose physics-informed loss terms as in physics-informed neural-networks (PINNs) \cite{Raissi2019}.
In our model, state variables are included in the training dataset together with design variables, so that minimizing $L_{\mathrm{recon}}$ automatically encourages the model to reproduce physically realistic circuit operational states.
In fact, \Cref{SubSectionResultsFidelityVerification} shows that the generated state variables agree well with SPICE simulations on average, meaning that the model naturally captures the underlying physics of the circuit operations.

There are several limitations of the current form of our model.
First, our model is not a direct feed-forward surrogate model: it does not output state variables from design variable inputs alone, or vice versa, because both are treated symmetrically as inputs/outputs of the VAE.
Conditioning on given design variables is instead possible through LSO with equality (`==') constraints on the design variables, although this requires a search rather than a single forward pass.
In this work, LSO uses gradient-free search strategies; gradient-based search through the differentiable decoder, as in the gradient-based optimization of FALCON \cite{Mehradfar2025}, is also possible in principle but is left for future work.
Next, our model does not currently provide topology generation capability as in CktGNN, CktGen, and OPAMP-Generator.
While it may be possible to include topological information as well as design/state variables as input/output of VAE to build higher-hierarchical latent space, such extension is not the near-future scope of our model.

Our near-future extensions might be the following: (a) topology-specific hyperparameter fine-tuning for each circuit topology, (b) application to larger circuits such as two-stage operational amplifiers, (c) testing the cross-topology transfer capability from one CMOS process technology node to another, (d) physical explainability evaluation based on latent space analysis, (e) integration with large-language-models to realize AMS CMOS circuit designer co-pilot systems, and (f) evaluation of LSO under a range of tighter specification requirements, to assess its controllability and the relative advantages of the search strategies, and (g) extension to AC and transient characteristics such as gain bandwidth and slew rate, by including capacitive state variables and additional simulation types.

\section{Conclusion}\label{SectionConclusion}

In this work, we proposed CktHGVAE: a heterogeneous graph variational autoencoder which takes not only design variables but also state variables as its inputs and outputs.
For a selection of transistor-level analog/mixed-signal CMOS circuits, its capabilities were demonstrated through closed-loop fidelity verification, latent-space optimization, and zero-shot cross-topology transfer.
Even with a single hyperparameter configuration without topology-specific tuning, comparison between model-generated and SPICE-simulated circuit operational states gave median $R^2$ values well above 0.9 for typical single-ended voltage amplifiers and the CMOS inverter with approximately 1,000 SPICE simulation samples, and 0.8141 - 0.9561 for differential pairs with approximately 2,000 samples.
Latent-space optimization in the PCA-reduced latent space with four major search strategies (Bayesian optimization, evolutionary search, random search, and quasi-exhaustive search) was demonstrated for a common-source voltage amplifier with NMOS input and resistive load (T1: CSA-N-Res).
For each search strategy, 28-30 of the 30 suggested design candidates satisfied the required circuit specifications at the decoder output, and SPICE simulations verified that 27-30 of them indeed satisfied the requirements.
It was also demonstrated that the model pretrained on a batch of topologies can be transferred to unseen topologies without retraining at all, although the transfer fidelity depended on the choice of pretraining topologies.
The model pretrained on single-ended topologies reconstructed differential-pair voltage amplifiers with zero-shot reconstruction median $R^2$ values of 0.720-0.901, while the fidelity was poor for one of the four scenarios.
All of these results suggest that a simple GVAE treating design and state variables symmetrically can serve as a basis for AMS CMOS circuit design automation.

\bibliographystyle{unsrtnat}
\bibliography{references}

\appendix

\section{Hyperparameters}\crefalias{section}{appendix}\label{AppendixHyperparameters}

\begin{table}[htbp]
  \caption{Model hyperparameters and configuration options.}
  \label{TableHyperparameters}
  \small
  \begin{tabularx}{\linewidth}{l>{\raggedright\arraybackslash}p{0.33\linewidth}X}
    \toprule
    Category & Setting & Value \\
    \midrule
    Architecture & Hidden representation width & 128 \\
    Architecture & Message-passing rounds & 3 \\
    Architecture & Activation function & SiLU \\
    Architecture & Dropout & 0.0 \\
    Architecture & Latent-vector size, device nodes & 4 \\
    Architecture & Latent-vector size, terminal pin nodes & 2 \\
    Architecture & Latent-vector size, net nodes & 2 \\
    Architecture & Node-level latent combination & Concatenation of per-node vectors \\
    Training & Optimizer & Adam \\
    Training & Learning rate & 0.001 \\
    Training & Batch size & 64 circuits per batch \\
    Training & Training epochs & 500 \\
    Training & Gradient-norm clipping & 50.0, active only after epoch 200 \\
    Loss & Reconstruction loss & Mean squared error, normalized per node type by its feature count \\
    Loss & KL-divergence weight schedule & Linear anneal from 0.0 to 0.01 over the first 50 epochs \\
    Loss & KL-divergence weight by node type & Device 1.0, pin 0.5, net 0.5 (relative to the annealed coefficient above) \\
    Latent-density fitting & Dimensionality reduction & Principal component analysis, retaining components explaining 95\% of variance \\
    Latent-density fitting & Density model selection & Gaussian mixture, number of components chosen by Bayesian information criterion \\
    \bottomrule
  \end{tabularx}
\end{table}

\end{document}